\documentclass[runningheads]{llncs}

\usepackage{eccv}

\usepackage{eccvabbrv}

\usepackage{graphicx}
\usepackage{booktabs}

\usepackage[accsupp]{axessibility}  

\usepackage{hyperref}
\usepackage{float} 

\usepackage{amsmath}
\usepackage{pifont}
\usepackage{float}  
\usepackage{makecell}
\usepackage{colortbl}
\usepackage{booktabs}
\usepackage{graphicx} 
\usepackage{CJKutf8}
\definecolor{lightyellow}{HTML}{FFF2CC}

\usepackage{multirow}
\usepackage{mathrsfs}
\usepackage{amssymb}
\usepackage{xcolor,amssymb,tikz}
\usepackage{amsmath}
\usepackage{tcolorbox}
\usepackage{listings}
\usepackage{orcidlink}

\begin{document}

\title{Global-Local Dual Perception for MLLMs in High-Resolution Text-Rich Image Translation} 

\titlerunning{GloTran}

\author{Junxin Lu\inst{1}\orcidlink{0000-0003-3653-0858} \and
Tengfei Song\inst{2} \and
Zhanglin Wu\inst{2} \and
Pengfei Li\inst{2}  \and
Xiaowei Liang\inst{2}  \and
Hui Yang\inst{2}  \and
Kun Chen\inst{2} \and
Ning Xie\inst{2} \and  
Yunfei Lu\inst{2} \and  
Jing Zhao \inst{3}\orcidlink{0000-0003-0158-5330}  \and
Shiliang Sun\inst{4}\orcidlink{0000-0001-7069-3752} \and
Daimeng Wei\inst{2} 
}

\authorrunning{Junxin Lu et al.}

\institute{School of Communication and Information Engineering, Shanghai University, Shanghai 200444, China \\
\email{junxinlu@shu.edu.cn}\\
\and
Huawei Translation Services Center, China \\
\email{songtf00@gmail.com, lipengfei203@h-partners.com \\ \{luyunfeicx,moonlight-007\}@163.com, \\ \{wuzhanglin2,liangxiaowei2,yanghui141,chenkun49,nicolas.xie\}@huawei.com}  
\and 
School of Computer Science and Technology, East China Normal University, Shanghai 200062, China \\
\email{jzhao2011@gmail.com} 
\and
State Key Laboratory of Submarine Geoscience, School of Automation and Intelligent Sensing, Shanghai Jiao Tong University, Shanghai 200240, China \\
\email{shiliangsun@gmail.com}
\thanks{Corresponding author: Shiliang Sun}}

\maketitle

\begin{abstract}
Text Image Machine Translation (TIMT) aims to translate text embedded in images in the source-language into target-language, 
requiring synergistic integration of visual perception and linguistic understanding.
Existing TIMT methods, whether cascaded pipelines or end-to-end multimodal large language models (MLLMs), 
struggle with high-resolution text-rich images due to cluttered layouts, diverse fonts, and non-textual distractions, resulting in text omission, semantic drift, and contextual inconsistency.
To address these challenges, we propose GLoTran, a global-local dual visual perception framework
for MLLM-based TIMT. 
GLoTran integrates a low-resolution global image with multi-scale region-level text image slices 
through pre-local region translation replay and an instruction-guided alignment strategy, 
conditioning MLLMs to maintain scene-level contextual consistency while faithfully capturing fine-grained textual details.
Moreover, to realize this dual-perception paradigm,  we construct GLoD, 
a large-scale text-rich TIMT dataset comprising 510K high-resolution global-local image-text pairs covering diverse real-world scenarios.
Extensive experiments demonstrate that GLoTran substantially improves translation completeness and accuracy over state-of-the-art MLLMs.
\end{abstract}

\begin{figure}[t]
\centering
\includegraphics[width=1\linewidth]{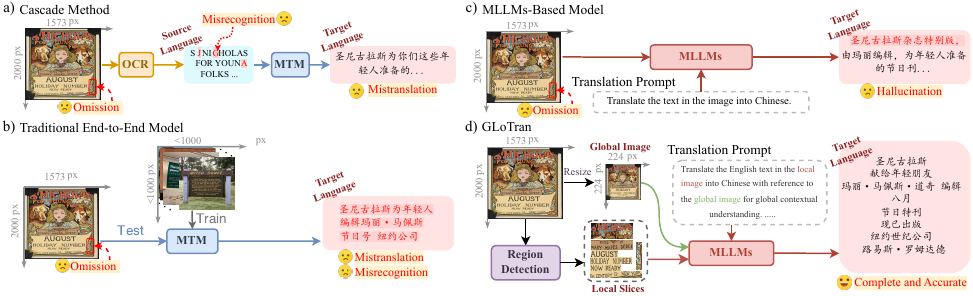}
\caption{Comparison of (a) cascade methods, (b) traditional end-to-end models, (c) MLLMs-based models, and (d) our proposed GLoTran for TIMT.
Through a dual visual perception strategy integrating global contextual understanding and local textual focus, 
we enable more complete and accurate TIMT. }
\label{intro}
\end{figure}

\section{Introduction}
Text image machine translation (TIMT) \cite{niu2024umtit,tian2023image} aims to translate source-language text embedded in images into a target-language, 
requiring a synergistic integration of visual perception and semantic understanding. 
Unlike conventional machine translation, which operates solely on text sequences,
TIMT simultaneously handles text detection, recognition, and translation within complex visual contexts.
Existing works primarily follow two paradigms: \emph{cascade methods} \cite{zhang2023novel,blecher2023nougat,hinami2021towards},
sequentially integrate optical character recognition (OCR) with neural machine translation,
and \emph{end-to-end models} \cite{ma2023multi,mansimov2020towards,zhang2023layoutdit,lan2024translatotron}, 
which jointly encode visual and linguistic modalities to directly map images into translated text. 
While both paradigms have driven progress, cascade methods suffer from error propagation and high computational cost, as illustrated in Fig.\ref{intro}(a), whereas conventional end-to-end models remain task-specific and struggle to generalize to diverse real-world scenarios, as shown in Fig.\ref{intro}(b).

Recently, multimodal large language models (MLLMs) have offered new avenues for TIMT.
By integrating strong language priors with visual perception, MLLMs provide a unified framework for multimodal understanding,  
achieving promising results in image comprehension \cite{park2024hierarchical,fang2025recognition,luan2024textcot}, 
document analysis \cite{ye2023ureader,zhang2025chaotic}, and visual question answering \cite{vu2025describe,fang2025guided,zhao2024lova3}.
Inspired by these capabilities, recent studies have adapted MLLMs for TIMT through prompting and instruction tuning 
\cite{zhang2024paying,zuo2025inimagetrans,liang2025single,gao2025multimodal}.   
Despite impressive progress, current MLLM-based TIMT methods still struggle with high-resolution text-rich images.
Real-world sources such as posters, screenshots, and documents often contain dense text interleaved 
with cluttered backgrounds, icons, and decorative elements, introducing non-textual distractions and producing 
an excessive number of visual tokens.
This redundancy disperses attention from meaningful text regions, 
frequently causing local text omission, misrecognition, and fragmented semantic understanding, 
as illustrated in Fig.\ref{intro}(c).
Furthermore, irregular layouts, heterogeneous fonts, and handwritten text disrupt spatial reasoning 
and compromise global-local semantic misalignment,
resulting in  translation omissions and hallucinations.
        
To address these limitations, we propose \textbf{GLoTran}, a global-local dual visual perception framework  for TIMT. 
GLoTran equips MLLMs with the ability to capture global context while preserving fine-grained focus on text-intensive regions,
effectively mitigating semantic drift, hallucinations, and omissions in visually complex images.
Specifically, we first perform text region detection to extract local visual slices, 
while downsampling the original high-resolution image to a low-resolution global view for efficient context encoding.   
Both the global image and local slices are jointly fed into the MLLM,
guided by an structured prompt that instructs the global image as a contextual reference and 
subsequent slices as its regional counterparts for per-slice translation.
The global image provides scene-level layout and semantic priors, ensuring contextual consistency,
while each local slice preserves detailed textual information critical for accurate text localization and translation, 
as shown in Fig.\ref{intro}(d).
Furthermore, we construct \textbf{GLoD}, a large-scale TIMT dataset tailed for our global-local dual perception paradigm. 
GLoD contains over 510K global-local image-text pairs from diverse real-world scenarios, 
including menu, documents, photos, posters, and road signs, facilitating effective fine-tuning of MLLMs for high-resolution text-rich TIMT.
Our main contributions can be summarized as follows:
\begin{itemize}
    \item We propose GLoTran, a global-local dual visual perception framework for MLLMs in TIMT, 
    which jointly models scene-level context understanding and fine-grained text regions perception, enabling accurate translation of high-resolution text-rich images.
    \item We construct a large-scale TIMT dataset with  over 510K global-local image-text pairs,
    systematically curated to facilitate global-local dual perception training of MLLMs.  
    \item Experimental results on multiple benchmarks demonstrate that GLoTran outperforms state-of-the-art MLLMs, 
    achieving higher translation accuracy for high-resolution text-rich TIMT.
\end{itemize}

\section{Related Work}
\label{Related_Work}
\textbf{\emph{Text Image Machine Translation.}} 
Different from  conventional  text machine translation \cite{yanglanguage,zhao2021knowledge,castilho2025survey,li2024non}, 
text image machine translation (TIMT) aims to translate the source-language texts embedded in images into a target-language
\cite{mansimov2020towards,lan2024translatotron}. 
Existing TIMT works can be broadly categorized into two paradigms: 
(\emph{i}) \emph{Cascade methods} sequentially perform text recognition and translation using multiple models, 
but are prone to error accumulation, structural redundancy, and high latency \cite{zhang2023novel,blecher2023nougat,hinami2021towards}, 
and (\emph{ii}) \emph{End-to-end models} jointly model visual and linguistic modalities to directly translate images to text, 
mitigating error propagation, and improving efficiency \cite{su2021rtnet,ma2023modal,ma2022improving,lan2024translatotron,niu2024umtit}. 
However,  cascade methods are inherently vulnerable to compound recognition and translation errors, while 
end-to-end models remain task-specific and lack the scalability and adaptability required for real-world scenarios
\cite{zhang2024paying,ma2023modal,zuo2025inimagetrans}.

\textbf{\emph{MLLM-Based TIMT.}}
Multimodal Large Language Models (MLLMs), such as  Qwen3-VL \cite{yang2025qwen3} and InternVL3 \cite{zhu2025internvl3},
have demonstrated the potential for end-to-end image-to-text translation.
Inspired by these advances, some works have explored adapting MLLMs more effectively for TIMT 
\cite{gao2024towards,zuo2025inimagetrans,liang2025single,liang2025improving,qian2024anytrans}.
MT$^3$ \cite{feng2025mt} improves the translation quality of MLLM through multi-task reinforcement learning, while M4Doc \cite{liang2025single}  focuses on MLLM-based document image translation.
InImageTrans \cite{zuo2025inimagetrans} introduces a multi-condition direct preference optimization strategy \cite{rafailov2023direct} to fine-tune MLLMs, 
reducing hallucinations and improving translation performance.
Despite their comparative performance, existing MLLM-based TIMT methods remain
 largely limited to low-resolution images with relatively simple visual content.
The application of MLLMs to end-to-end TIMT for real-world high-resolution text-rich images is still underexplored.

For high-resolution images, although some works \cite{park2024hierarchical,wu2024v,wang2025divide,shen2025zoomeye,cailocal} 
combine global views with local patches, 
they primarily leverage local patches as auxiliary 
visual evidence for global queries to support holistic VQA and reasoning tasks.
Moreover, their random partitioning \cite{park2024hierarchical,shen2025zoomeye,wang2025divide} or search-based strategies 
\cite{cailocal,wu2024v} often lead to fragmented text, 
redundant regions, and missing text instances.
However, TIMT requires comprehensive, ordered, and semantically consistent translation of all textual content in an image,
rendering they not directly applicable.
High-resolution images, such as menus and posters, typically contain cluttered backgrounds, 
diverse layouts, and non-textual visual distractions. 
These characteristics make it challenging for existing models to maintain fine-grained textual focus 
and global contextual consistency, resulting in hallucinations, semantic drift, and cross-region inconsistencies.
In contrast, GLoTran adopts a text-region-centered extraction and perception paradigm.  
It translates each local region while actively querying global context for relevant semantic cues 
(e.g., layout and theme), and enforces translation coherence through pre-local region translation replay and 
a structured global-local prompting strategy.

\begin{figure*}[t]
\centering
\includegraphics[width=0.99\linewidth]{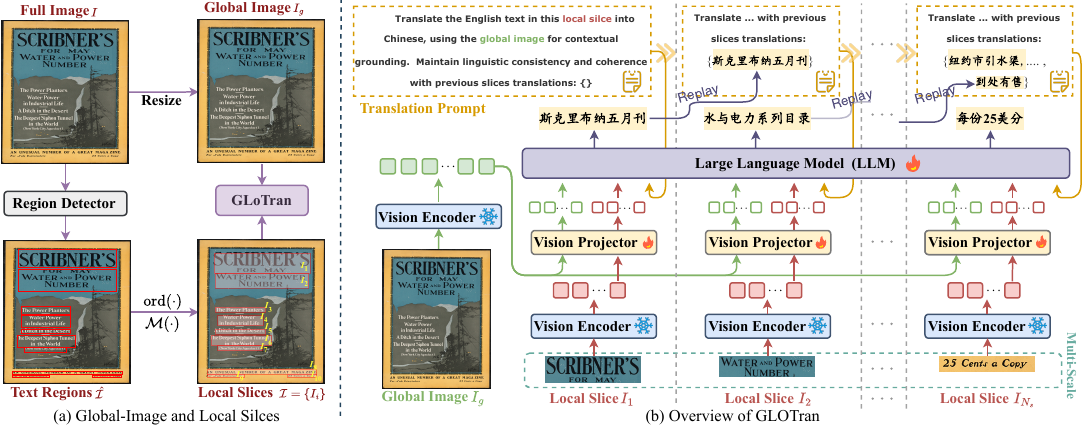}
\caption{Overview of the proposed GLoTran framework. 
(a) The high-resolution input image is processed by a text region detector to identify candidate textual regions, which are subsequently sorted, merged, and cropped into localized slices. 
(b)  The global image and the local slices are fed into GLoTran with a structured  prompt, enabling global contextual understanding and local textual focus for TIMT.}
\label{overview}
\end{figure*}

\section{Methodology}
\label{Method}
In this section, we introduce GLoTran, an efficient and scalable framework tailored for TIMT. 
GLoTran adopts a global-local dual visual perception strategy, 
enabling comprehensive contextual understanding while maintaining fine-grained attention to local textual regions.
 
\subsection{Architecture Overview} 
TIMT translating source-language text embedded within an image into a target-language sequence. 
Formally, given a TIMT dataset $\mathcal{D}=\{(I, Y)\}$, where $I$ denotes a source-language image
and $Y$ represents its corresponding target-language translation. 
TIMT seeks to learn a translator $\mathcal{F}:I\rightarrow Y$, such that the predicted sequence $\hat{Y}=\mathcal{F}(I)$
faithfully preserves the semantics of $Y$. 

Unlike existing MLLM-based methods that perform translation over an entire high-resolution image in a single forward pass,
GLoTran adopts a global-local dual visual perception strategy to balance holistic contextual understanding with fine-grained textual focus.
As illustrated in Fig.\ref{overview}, GLoTran first detects translatable text regions within $I$  
and extracts the cropped local slices $\mathcal{I}=\{I_1,I_2,\cdots,I_{N_s}\}$. 
The original image $I$ is simultaneously resized into a low-resolution global view $I_g$ that captures the overall layout and semantics of the scene, 
while each local slice $I_i$ preserves detailed text content for fine-grained translation.
The translation is then performed regressively across regions, guided by a structured instruction prompt $\mathcal{P}$ (see Sec.~\ref{Prompt_Construction}).
At each translation step $i$, GLoTran predicts $\hat{Y}_i$ for $I_i$,
conditioned on $I_g$, $I_i$ and $\mathcal{P}_{<i}$.
$\mathcal{P}_{<i}$ incorporates the translated outputs $\{\hat{Y}_{i-\eta}, \ldots, \hat{Y}_{i-1}\}$ from the preceding $\eta$ regions, 
 where $\eta$ is termed as pre-region translation replay window. 
This replay mechanism provides contextual cues such as sentence continuation, 
coreference grounding, and terminology consistency, thereby improving the accuracy and discourse coherence of local translations within complex, multi-text images.
 
\subsection{Global-Local Dual Visual Perception} 
TIMT requires both scene-level semantic understanding and fine-grained textual recognition.
However, pretrained MLLMs are constrained by limited input resolutions and the rigid square-patch design of their visual encoders.
When MLLMs directly process high-resolution text-rich images, the superlinear computational overhead and dense non-text visual distractions overload visual attention, thereby weakening fine-grained text recognition in TIMT and significantly degrading translation accuracy.
Consequently, we propose a global-local dual visual perception strategy that jointly models a low-resolution global view and multi-scale local slices,
achieving a balance between global contextual understanding and local textual fidelity.

Specifically, a self-trained  text region detector (see Appendix 10 for details) 
identifies candidate bounding boxes $\mathcal{B}=\{b_i\}_{i=1}^{N_b}$ from image $I$,
where each box corresponds to a block- or paragraph-level region that groups sentences from the same coherent text segment.
Each detected region is cropped and normalized to form an initial local slice set 
$\mathcal{\hat{I}}=\{\hat{I}_i=\text{Crop}(I, b_i)\}_{i=1}^{N_b}$.
Adjacent or semantically related regions are further merged according to geometric and typographic cues (e.g., overlap, line height, alignment),  
resulting in a compact grouped slice set $\mathcal{\tilde{I}}=\mathcal{M}(\mathcal{\hat{I}})=\{\tilde{I}_i\}_{i=1}^{N_s}$, with $N_s \ll N_b$. 
To ensure translation coherence, the slices are deterministically arranged using a content ordering model  (e.g., MinerU \cite{wang2024mineru})
$\text{ord}(\cdot)$, yielding $\mathcal{I}=\text{ord}(\mathcal{\tilde{I}})$. 
 Meanwhile, the original image $I$ is downsampled into a low-resolution global view $I_g$, 
which captures holistic scene layout and contextual priors, whereas each local slice $I_i$ retains fine-grained textual  details.

Both $I_g$ and $I_i$ are independently encoded by a shared visual encoder (e.g., ViT \cite{dosovitskiy2020image}) 
to extract visual features $\tilde{V}_g$ and $\tilde{V}_i$:
\begin{equation}
    \tilde{V}_g = \text{Enc}(I_g) \in \mathbb{R}^{n_g \times d_v},   \tilde{V}_i = \text{Enc}(I_i)  \in \mathbb{R}^{n_s  \times d_v},
\end{equation}
where $d_v$ denotes the visual feature dimension, and $n_g$ and $n_s$ correspond to the number of patch tokens for the global and local views, respectively.
Subsequently, a vision projector aligns visual features to the textual feature space, 
producing $V_g \in \mathbb{R}^{n_g \times d_t}$ and  $V_i \in \mathbb{R}^{n_s \times d_t}$, where $d_t$ denotes the text embedding dimension.
Overall, the decoding of the LLM for global-local dual visual input is defined as:
\begin{equation}
    \hat{Y}_i = \text{LLM}([E_g;V_g;E_l;V_i;\mathcal{P}_{<i}] \in \mathbb{R}^{L \times d_t}),
\end{equation}
where $E_g$ and $E_l$ are textual embeddings of source-aware identifier tokens $\left\langle\text{image}_g\right\rangle$ and 
$\left\langle\text{image}_l\right\rangle$, which help the LLM distinguish features from global and local views.

To effectively integrate global context into local perception while preserving local text details,
we introduce hierarchical cross-attention between $V_g$ and $V_i$ at early Transformer layers [0, 8, 16, 24].
This allows each local token to selectively attend to semantically relevant global tokens, 
thereby enhancing contextual grounding and maintaining coherence across the visual hierarchy  \cite{wu2025flashbias,yemplug}. 
Formally, for a query $q_i$ from a local token and a key $k_j$ from a global token, 
the attention weight is computed as: 
\begin{equation} 
\text{Attn}(q_i, k_j) = \frac{q_i k_j^\top}{\sqrt{d_t}} + b_{ij}, 
\end{equation} 
where $b_{ij}$ is a learnable scalar bias conditioned on source type (global vs. local) and token spatial proximity.
This adaptive bias encourages semantically grounded and spatially aware interactions, 
enabling model to prioritize relevant global cues for local disambiguation (e.g., resolving ambiguous text via scene context)
while preserving intra-line coherence and  ignoring irrelevant regions. 

\begin{figure}[t]
\centering
\includegraphics[width=1.0\linewidth]{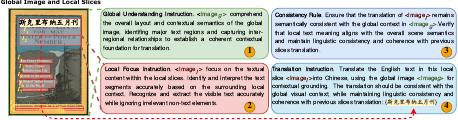}
\caption{An illustration of the structured prompt design in GLoTran.}
\label{prompt}
\end{figure}

The training objective of GLoTran, parameterized by $\theta$, is defined as:
\begin{equation}
\begin{aligned}
& \theta^* = \arg\min_{\theta} \mathcal{L}\big( \mathcal{F}(I, \mathcal{P}); \theta \big) \\
& = \arg\min_{\theta} \frac{1}{|\mathcal{D}|} 
\sum_{(I, Y) \in \mathcal{D}} 
\left[- \sum_{i=1}^{N_s} Y_i \log p(\hat{Y}_i \mid I_g, I_i, \mathcal{P}_{<i}; \theta) \right],
\end{aligned}
\end{equation}
where $\mathcal{L}$ denotes the loss of cross-entropy on target-language tokens.
During training, teacher forcing \cite{chen2024diffusion,tian2025beyond} is applied to replace previously predicted translations $\hat{Y}_{<i}$ with ground-truth $Y_{<i}$ within the replay window $\eta$, 
ensuring stable optimization and effective cross-region contextual learning.
 
\subsection{Prompt Construction of GLoTran}
\label{Prompt_Construction}
Within GLoTran, Translation is performed in a regressive regional manner, 
where each local translation step is conditioned on the global image, the current local slice,   
and previously regions translations through a replay mechanism, thereby ensuring translation  consistency and coherence  across regions.
As illustrated in Fig.\ref{prompt}, the structured prompt $\mathcal{P}$ is composed of four interdependent components:
\begin{itemize} 
   \item \textbf{Global Understanding Instruction} encodes holistic scene semantics and spatial layout from the global image, 
   establishing a contextual foundation for subsequent region-level translations.
    \item  \textbf{Local Focus Instruction} emphasizes the active text region in current local slice, 
    steering MLLMs attention toward fine-grained textual understanding while preserving alignment with the global contextual and visual layout.
    \item  \textbf{Global-Local Consistency Rule} enforces semantic coherence between regional translations and the global view.
Before generating outputs, MLLM validates that local translations are consistent with global scene semantics, mitigating contextual drift and contextual deviation.
Furthermore, translated outputs from preceding regions $\{\hat{Y}_{i-\eta}, \ldots, \hat{Y}_{i-1}\}$ are incorporated into the prompt via a translation replay window of size $\eta$, providing discourse cues for terminology uniformity and narrative continuity across slices.
     \item  \textbf{Translation Instruction} specifies the explicit TIMT objective, 
     e.g., {\itshape Translate the English text in this local slice into Chinese, using the global image  for contextual grounding.
     Maintain linguistic consistency and coherence with previous slices translation:\{\colorbox{lightyellow}{\footnotesize\begin{CJK}{UTF8}{gbsn}斯克里布纳五月刊\end{CJK}\}}}
     is provided when translating the second slice $I_2$, where the previous translation 
     \colorbox{lightyellow}{{\itshape\footnotesize\begin{CJK}{UTF8}{gbsn}斯克里布纳五月刊\end{CJK}}} originates from the slice $I_1$.
     This explicit directive reinforces task fidelity and maintains linguistic fluency across regional decoding steps.
\end{itemize}
 
\section{GLoD Dataset}
\label{dataset}
\subsection{Dataset Statistics}
\begin{figure}[]
\centering
\includegraphics[width=1.0\linewidth]{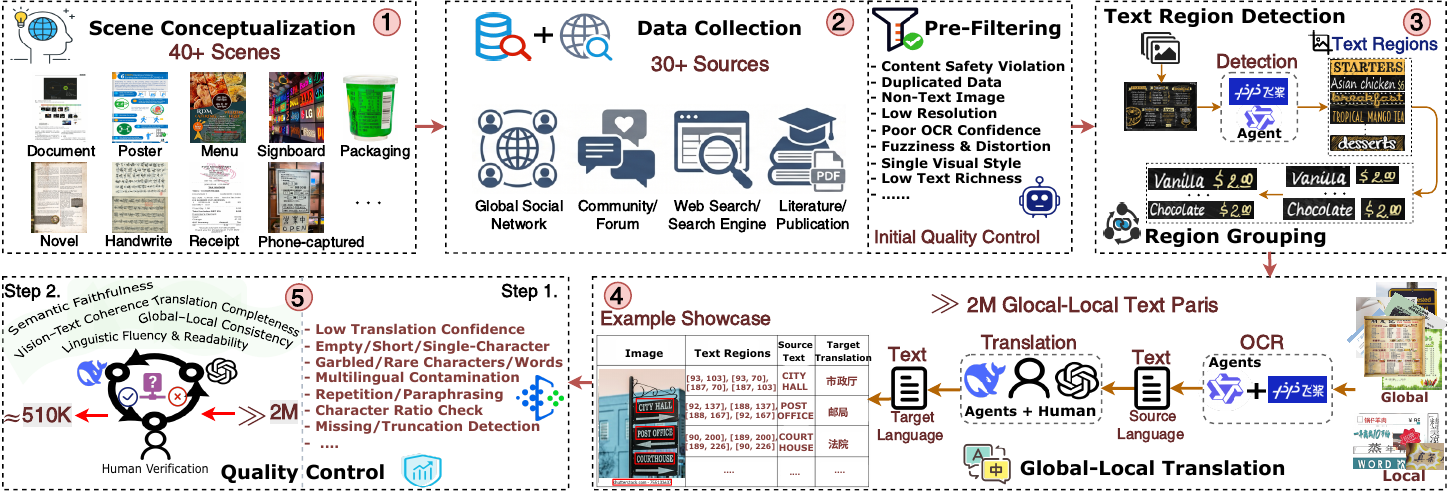}
\caption{Overview of GLoD curation pipeline. The construction pipeline systematically generates high-quality global and local translation pairs through five core stages: 
conceptualization, data collection and pre-filtering, 
text region detection and grouping, 
global-local translation, and quality control.}
\label{dataset_curation}
\end{figure}

Tab.\ref{statistics_dataset} provides a comparative overview of our GLoD dataset against existing TIMT datasets, 
including OCRMT30K \cite{lan2023exploring}, DoTA \cite{liang2024document}, IIMT \cite{lan2024translatotron}, and MIT-10M \cite{li202510m}.
While most prior datasets offer full-image-level pairs, where each image aligns with a single holistic translation, such coarse-grained structures limit training for our global-local dual visual perception frameworks.
To address this, we construct GLoD, a large-scale multimodal dataset tailored for global-local dual-input TIMT in text-rich scenarios,
comprising 510K images and 517,354 image-text pairs, spanning over 40 real-world scenes in 5 languages.
Curated via a rigorous pipeline, it features high-resolution, text-rich content, irregular layouts, complex backgrounds, and diverse fonts, as shown in Fig.\ref{dataset_curation}.

\begin{table}[]
\centering
\caption{The statistics of GLoD and other TIMT dataset.}
\resizebox{0.8\columnwidth}{!}{
\begin{tabular}{cccccc}
\toprule 
\rowcolor{gray!20}   \textbf{Dataset}  &  \textbf{Text-Rich} & \textbf{Global-Local} & \textbf{Language} & \textbf{Images} &   \textbf{Image-Text} \\ \midrule
ECOIT    \cite{zhu2023peit}             & \ding{55}      & \ding{55}   & 2   & 479,490 & 479,490  \\
OCRMT30K \cite{lan2023exploring}        & \ding{55}      & \ding{55}   & 2   & 30,000  & 30,000 \\
DoTA     \cite{liang2024document}       & \ding{51}      & \ding{55}   & 2   & 126,345 & 126,000 \\
IIMT     \cite{lan2024translatotron}    & \ding{55}      & \ding{55}   & 1   & 89,033  & 89,033 \\
MIT-10M  \cite{li202510m}               & \ding{55}      & \ding{55}   & 14  & 840,855 & 10,931,115 \\ \midrule
\rowcolor[HTML]{E3F2FD} GLoD     & \ding{51}  & \ding{51}  & 5 &  517,354  & 517,354  \\     
\bottomrule
\end{tabular}}
\label{statistics_dataset}
\end{table}

\subsection{Curation Pipeline of GLoD}
The GLoD curation pipeline encompasses five key stages: scene conceptualization, data collection and pre-filtering, 
text region detection and grouping, global-local translation, and quality control.

\textbf{Scene Conceptualization.} We define over 40 practical translation scenarios, 
including documents, posters, menus, road signs, and receipts, ensuring broad coverage of visual and content diversity. 


\textbf{Data Collection and Pre-Filtering.} Images are collected from publicly available Internet resources and integrated with existing text-image datasets.
Strict filters exclude privacy-sensitive, confidential, or copyright-protected content, 
as well as low-resolution, blurred, distorted, or poor OCR confidence (using PaddleOCR \cite{cui2025paddleocr}), uniform visual styles, 
and low text richness images.

\textbf{Text Region Detection and Grouping.} 
Filtered images undergo automated text region detection using complementary detectors, 
PaddleOCR and Qwen3-VL-Plus \cite{yang2025qwen3}, 
to capture diverse text layouts through cross-region verification and complementary fusion.
Detected regions are hierarchically grouped into semantically coherent blocks (phrases, sentences, or paragraphs) based on reading order, spatial adjacency, and visual heuristics, while ambiguous or underspecified regions are excluded.

\textbf{Global-Local Translation.} 
Each global–local pair is processed through dual OCR recognition and bi-directional translation fusion. 
PaddleOCR captures fine-grained local content, while Qwen3-VL-Plus provides global context for cross-line disambiguation, enabling robust recognition 
under diverse fonts, print qualities, and image sizes. 
The recognized text is then translated using GPT-4o \cite{hurst2024gpt} and DeepSeek-R1 \cite{guo2025deepseek} under a multi-round mutual back-translation and fusion process, ensuring linguistic completeness and contextual alignment. Human experts conduct final verification and correction, focusing on semantics, grammar, and layout correspondence. This pipeline produces approximately 2M global–local image-text pairs with high linguistic and structural fidelity.

\textbf{Quality Control.} 
To guaranty reliability, we employ a two-stage validation strategy.
Low-confidence OCR segments, ultra-short translations, and non-linguistic symbols are first removed.
Subsequently, semantic faithfulness and consistency between global and local translations are quantified using multilingual embedding similarity 
(e.g., LaBSE \cite{feng2020language} or SimCSE \cite{gao2021simcse}) and round-trip translation consistency. Local regions showing semantic drift or contextual mismatch with their global references are excluded.

\begin{table*}[!t]
\centering
\caption{Performance comparison of translating English into Chinese with open-source and close-source MLLMs on the MCiTon \cite{zuo2025inimagetrans} test set. 
We report \textbf{BLEU/COMET} for translation quality of each real-world scenarios. 
$\uparrow$  indicates higher is better.  
The \textbf{bold} represents the best results, and the \underline{underline} represents the second best results.}
\resizebox{\textwidth}{!}{
\begin{tabular}{r|r|ccc|ccc|cc|c}
\toprule 
& & \multicolumn{3}{c|}{\textbf{Document}} & \multicolumn{3}{c|}{\textbf{Scene}} & \multicolumn{2}{c|}{\textbf{Poster}} \\ \cmidrule(lr){3-5} \cmidrule(lr){6-8} \cmidrule(lr){9-10}
\multirow{-2}{*}{\textbf{Method}} & \multirow{-2}{*}{\textbf{Size}} & \textbf{Paper} $\uparrow$ & \textbf{News} $\uparrow$& \textbf{Novel} $\uparrow$ & \textbf{Title} $\uparrow$ & \textbf{Sign} $\uparrow$ & \textbf{Introduction} $\uparrow$ & \textbf{Cover} $\uparrow$ & \textbf{Leaflet} $\uparrow$ & \multirow{-2}{*}{\textbf{Avg}$\uparrow$}\\
\midrule 
\rowcolor{gray!20}  \multicolumn{11}{c}{\textbf{Close-Source   MLLMs}} \\ \midrule 
GPT-4o              \cite{hurst2024gpt}       & $\thicksim$ 200B  & 62.1 / 86.3  & 51.6 / 85.6  & 44.2  /  83.4 & 51.3 / 83.4 &  46.1 / 83.3  &  40.1 / 75.3 & 34.9 / 69.1  &  43.9 / 71.7 & 46.8 / 79.9  \\
Qwen-VL-Max         \cite{bai2023qwen}        & $\thicksim$ 72B  & 60.5 / 85.7  & 58.3 / 85.7  & 44.8  /  83.5 & 50.1 / 81.6 &  43.2 / 80.5  &  37.8 / 74.6 & 34.2 / 67.3  &  43.7 / 69.9 & 46.5 / 78.6  \\ \midrule
\rowcolor{gray!20}  \multicolumn{11}{c}{\textbf{Open-Source   MLLMs}} \\ \midrule
Qwen2.5-VL-Instruct  \cite{bai2025qwen2}      & 7B  & 60.7 / 85.8  & 53.4 / 85.5  & 27.4  / 78.1 & 47.9 / 80.1 &  44.0 / 82.2  &  36.2 / 72.8 & 30.3 / 64.4  &  43.6 / 72.4  & 42.9 / 77.7  \\
Qwen2-VL             \cite{wang2024qwen2}     & 7B  & 60.0 / 85.6  & 46.4 / 81.7  & 31.8  / 73.9 & 27.9 / 67.8 &  25.9 / 66.4  &  24.0 / 70.9 & 19.4 / 70.3  &  30.4 / 73.3  & 36.9 / 73.1 \\
DeepSeek-VL-Chat     \cite{lu2024deepseek}    & 7B  & 44.7 / 80.8  & 8.7 / 57.4   & 4.9 / 50.6   & 5.4 / 47.6  &  13.5 / 59.9  &  10.8 / 52.1 & 9.7 / 45.8   & 13.3 / 52.9   & 13.9 / 55.9 \\
InternVL2           \cite{chen2024internvl}   & 8B  & 44.0 / 83.0  & 35.8 / 78.9  & 23.7  / 71.3 & 27.2 / 65.3 &  28.2 / 61.1  &  25.4 / 68.7 & 19.4 / 62.0  &  21.0 / 62.1  & 28.1 / 68.5  \\
InImageTrans        \cite{zuo2025inimagetrans}& 8B  & 59.0 / 85.0  & 46.5 / 80.9  & 33.5  / 76.5 & 37.1 / 70.6 &  28.9 / 66.9  &  32.0 / 71.7 & 19.0 / 62.3  &  32.1 / 70.8  & 35.9 / 72.7  \\
InternVL3           \cite{zhu2025internvl3}   & 8B  & 59.8 / 86.0  & 48.9 / 83.9  & 31.2  / 79.8 & 50.5 / 82.1 &  41.8 / 81.6  &  32.8 / 72.6 & 29.5 / 66.1  &  35.8 / 70.9  & 41.4 / 77.9  \\
Qianfan-VL          \cite{dong2025qianfan}    & 8B  & 55.4 / 83.3  & 52.9 / 81.2  & 32.0  / 80.7 & 47.5 / 78.4 &  43.0 / 80.9  &  34.4 / 71.7 & 25.1 / 61.7  &  34.5 / 69.3  & 40.6 / 75.9 \\
SAIL-VL2            \cite{dong2025scalable}   & 8B  & 61.3 / 85.4  & 53.6 / 84.9  & 43.4 / 81.6  & 48.8 / 78.8 &  43.5 /  82.1 &  36.9 / 73.5 & 32.9 / 66.4  & 40.1 / 71.5   & 45.1 / 78.0\\
Qwen3-VL-Instruct   \cite{yang2025qwen3}      & 8B  & 60.9 / 86.1  & 54.3 /  \underline{86.2}  &  \underline{43.6}  /  \underline{83.1} &  \underline{52.3} / 81.8 &  44.2 /  \underline{83.0}  &  33.6 /  \underline{75.8} &  \underline{36.2} /  \underline{70.4}  &   \underline{48.2} /  \underline{75.1}  & 46.7 / 80.2  \\  
InternLM-XComposer2 \cite{dong2024internlm}   & 11B & 37.3 / 78.4  & 26.7 / 60.1  & 8.50  / 45.2 & 24.1 / 60.2 &  22.4 / 60.0  &  20.2 / 61.1 & 16.6 / 57.3  &  22.0 / 60.0  & 22.7 / 60.6 \\
InternVL3           \cite{zhu2025internvl3}   & 14B &  \underline{63.2} / \underline{86.4}  & 53.8 / 85.4  & 36.7  / 81.5 & 49.1 / 81.5 &  42.4 / 81.8  &  34.7 / 73.8 & 30.1 / 66.2  &  35.3 / 70.9  & 41.7 / 78.2 \\
CogVLM              \cite{wang2024cogvlm}     & 17B & 59.2 / 84.7  & 44.8 / 80.6  & 30.7  / 73.7 & 36.4 / 71.1 &  27.7 / 66.7  &  30.5 / 70.6 & 25.7 / 67.8  &  30.1 / 71.8  & 35.5 / 72.7  \\
Qwen2.5-VL-Instruct \cite{bai2025qwen2}       & 32B & \underline{63.2} / 86.3  &  \underline{54.4} /  \underline{86.2}  & 28.5  / 79.2 & 39.1 / 77.3 &   \underline{46.2} / 82.6  &   \underline{39.9} / 75.7 & 33.9 / 68.1  &  45.5 / 72.8  & 42.2 / 78.6 \\
InternVL2           \cite{chen2024internvl}   & 40B & 48.8 / 83.9  & 40.1 / 79.8  & 26.6  / 72.8 & 27.1 / 67.1 &  28.4 / 62.5  &  28.7 / 69.9 & 19.7 / 64.1  &  23.4 / 66.1  & 30.3 / 70.3  \\ 
\midrule
InternVL3$^\bigstar$\cite{zhu2025internvl3}          &  8B  & 58.3 / 86.3 &  50.5 / 85.5  &  32.1 / 81.4 &  52.4 / 83.4 &  38.1 / 82.3 &  34.6 / 73.6 &  31.0 / 68.5  & 33.8 / 69.3 & 41.4 / 78.8  \\   
Qwen3-VL-Instruct$^\bigstar$\cite{yang2025qwen3}     &  8B  & 60.8 / 86.2 &  56.0 / 86.4  &  43.8 / 83.9 &  46.1 / 80.5 &  44.7 / 83.8 &  35.5 / 77.4 &  36.2 / 72.6  & 48.5 / 75.9 & 46.5 / 80.8  \\
GLoTran-LoRA (Qwen3-VL 8B) & 8B & 64.8 / 85.9 & 56.0 / 86.7 & 45.5 / 84.8 & 53.1 / 81.2 & 45.0 / 82.7 & 40.0 / 74.5 & 36.1 / 69.9 & 49.5 / 75.8 & 42.2 / 78.6 \\
\rowcolor[HTML]{BBDEFB}  GLoTran (Qwen3-VL 8B)   & 8B  &  \textbf{66.3 / 87.9} &  \textbf{57.5 / 88.6}  & \textbf{46.8 / 86.6} &  \textbf{54.3 / 83.2}  &   \textbf{46.3 / 84.8}  & \textbf{41.3 / 76.4} &  \textbf{37.7 / 72.0}  & \textbf{51.1 / 77.9}  &  \textbf{50.2 / 82.2} \\
\bottomrule 
\end{tabular}
}
\label{MCiTon_results}
\end{table*}

\section{Experiments}
\label{Experiments}
\subsection{Training and Test datasets}
We conduct full-parameter fine-tuning of existing MLLMs, e.g., InternVL2.5 4B \cite{zhu2025internvl3} and Qwen3-VL 8B \cite{yang2025qwen3}, 
on GLOD using global-local dual visual perception strategy.
To ensure a fair comparison, we evaluate the English-to-Chinese translation performance on the MCiTon \cite{zuo2025inimagetrans} dataset, 
which contains 1,450 image-text pairs from documents, nature scenes and posters.
Additionally, we evaluate multilingual translation capabilities on the MTIT6 dataset \cite{qian2024anytrans}, 
which includes five translation tasks:  
Japanese-to-Chinese (jp$\rightarrow$zh), Korean-to-Chinese (ko$\rightarrow$zh), Chinese-to-English (zh$\rightarrow$en), 
Chinese-to-Japanese (zh$\rightarrow$jp) and Chinese-to-Korean (zh$\rightarrow$ko), with approximately 200 high-resolution images per language pair.
 
\subsection{Baselines and Implementation Details}
We compare GLoTran with a wide range of open-source MLLMs, including 
Qwen2.5-VL-Instruct  \cite{bai2025qwen2}, Qwen3-VL \cite{yang2025qwen3},  and InternVL3.
We also include comparisons with state-of-the-art commercial MLLMs, namely GPT-4o \cite{hurst2024gpt} and Qwen-VL-Max \cite{bai2023qwen}.
For evaluation of translation quality, we adopt the metrics BLUE \cite{papineni2002bleu} and COMET \cite{rei2020comet},
following the same evaluation procedures as InImageTrans \cite{zuo2025inimagetrans}. 
All reported results are obtained either from officially released benchmarks or 
via re-inference using publicly available checkpoints to ensure fairness and consistency.
GLoD contains 512,180 training pairs and 5,174 validation pairs under a 99:1 split, with no image overlap with MCiTon \cite{zuo2025inimagetrans} or MTIT6 \cite{qian2024anytrans}.
All original input images are resized to a fixed resolution of 224$\times224$. 
During full-parameter fine-tuning, the batch size is set to 128 with gradient accumulation over 8 steps. 
The AdamW optimizer is used with an initial learning rate of 1.0$\times$10$^{-5}$, and the model is trained for 10,000 iterations in total. 

\begin{figure}[tbp]
\centering
\includegraphics[width=0.8\linewidth]{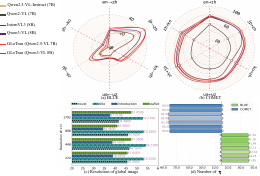}
\caption{(a)-(b) performance comparison of multilingual translating task with open-source MLLMs on MTIT6 \cite{qian2024anytrans}. 
(c)-(d) performance analysis with respect to  global image resolution and the pre-region translation replay window size $\eta$.}
\label{multilingual_results}
\end{figure}

\begin{figure}[t]
\centering
\includegraphics[width=1.0\linewidth]{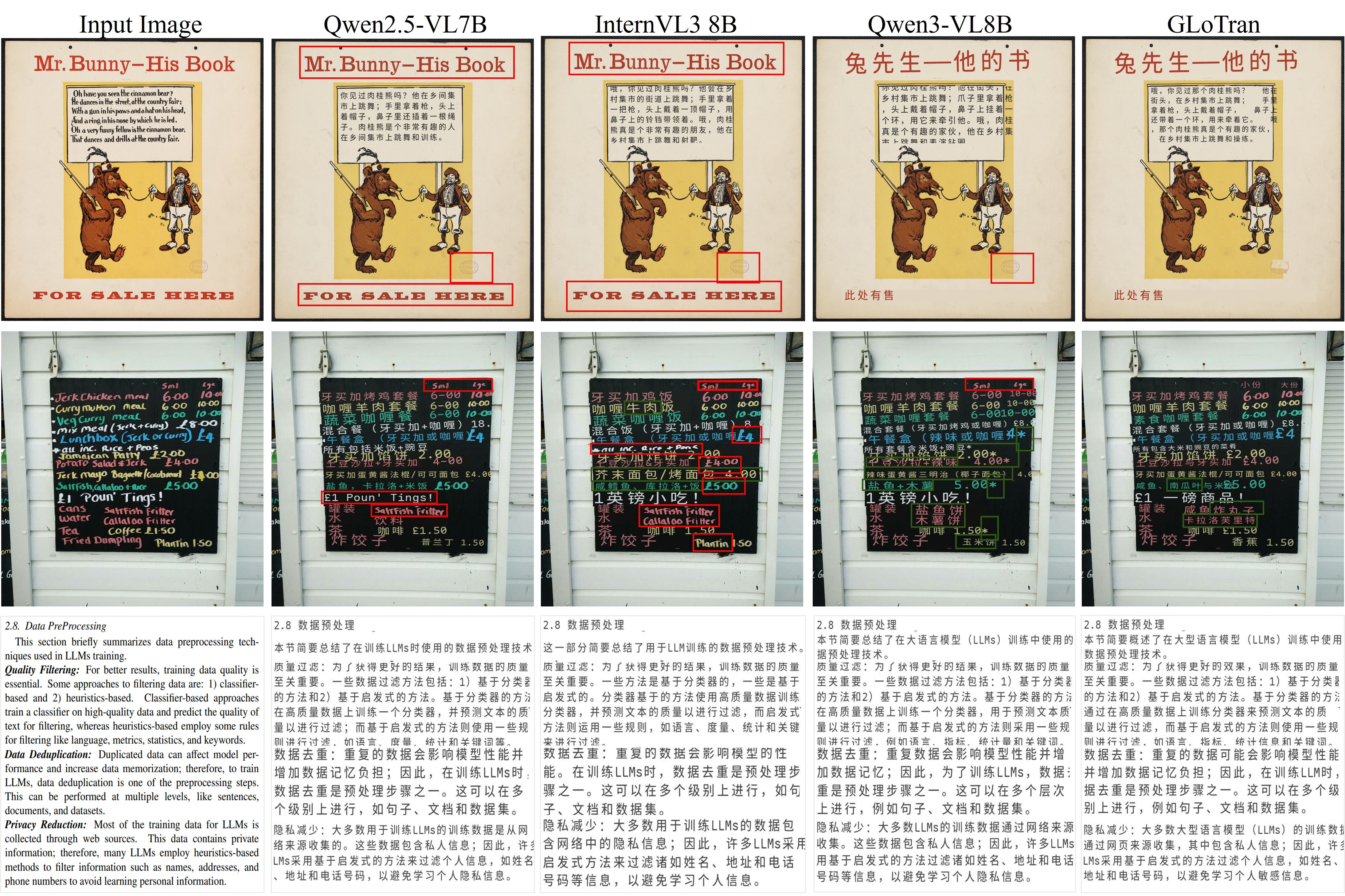}
\caption{Qualitative comparison. The results in \textcolor{red}{$\square$}  indicates omissions, those in \textcolor[HTML]{305918}{$\square$} indicate mistranslations, and those in \textcolor[HTML]{611150}{$\square$} indicate hallucinations caused by contextual misunderstanding.  
}
\label{Qualitative}
\end{figure}

\subsection{Performance Comparison}
\textbf{\emph{Multi-Scenes Translation Evaluation}.} 
Tab.\ref{MCiTon_results} reports the performance of various MLLMs on the MCiTon \cite{zuo2025inimagetrans} benchmark.
As shown in Tab.\ref{MCiTon_results}, GLoTran, built on Qwen3-VL 8B, achieves improvements across all eight scenarios,
outperforming most open-source and closed-source MLLMs of comparable or larger scales.
GLoTran surpasses Qwen3-VL 8B, achieving gains of 7.49$\%$ and 2.49$\%$ in BLEU and COMET,
highlighting the effectiveness of GLoTran in improving sensitivity to fine-grained textual details while preserving global contextual comprehension.
In dense and heterogeneous layouts (e.g., documents and posters), GLoTran improves BLEU by
4.6$\%$ and 5.4$\%$ average gains over all open-source MLLMs.
In scenes with small and scattered texts (e.g., introductions and leaflets), GLoTran achieves larger relative gains,
outperforming Qwen3-VL 8B by approximately 5.3 BLEU and 1.7 COMET points,
while achieving comparable improvements with Qwen2.5-VL-Instruct requires scaling its parameters from 7B to 32B.
We further perform supervised fine-tuning (SFT) on our GLoD dataset for InternVL3 8B and Qwen3-VL 8B, 
denoted as InternVL3$^\bigstar$ and Qwen3-VL$^\bigstar$ in Tab.\ref{MCiTon_results}. 
The results indicate that directly fine-tuning these models on TIMT with GLoD, 
without tailored adaptation for high-resolution text-rich image translation, 
brings only limited gains and may even degrade performance in complex scenarios.
These consistent improvements confirm that GLoTran effectively captures subtle textual variations in visually complex regions,
alleviating fine-grained translation errors that commonly challenge MLLMs.

Furthermore, the results in Tab.\ref{MCiTon_results} indicate that the performance of MLLMs in TIMT tasks does not exhibit a strong positive correlation with the parameter scale.
For models built on the same backbone, increasing the parameters beyond 7B yields only marginal or negligible gains.
Specifically, InternVL2 8B and InternVL2 40B differ by merely 2.2 BLEU and 1.8 COMET points (28.1/68.5 vs. 30.3/70.3). In contrast, Qwen2.5-VL-Instruct 32B even underperforms its 7B variant in BLEU (42.2 vs. 42.9).
These observations suggest a potential scaling saturation effect, which implies that the larger model capacity alone may not adequately address the intrinsic challenges of TIMT.

\textbf{\emph{Multilingual Translation Evaluation}.} 
To further evaluate cross-lingual performance, we conduct multilingual translation experiments on the MTIT6 \cite{qian2024anytrans} 
dataset, covering six language pairs. As shown in Fig.\ref{multilingual_results}(a)-(b), 
GLoTran consistently achieves the best performance in all tasks, 
surpassing Qwen3-VL 8B and InternVL3 8B. 
Although InternVL3 and Qwen3-VL demonstrate strong multilingual understanding and text-image translation, 
they are constrained by two critical limitations: (\emph{i}) insufficient suppression of non-textual visual distractions, 
which can produce hallucinatory or contextually inconsistent translations, and
(\emph{ii}) inadequate attention to small or fine-grained textual regions, 
leading to omissions or mistranslations of marginal or low-contrast text. 
These limitations contribute to the performance plateau observed in existing MLLMs, even as their parameter counts increase, underscoring the need for strategies that jointly capture global contextual coherence and fine-grained local perception.

\textbf{\emph{Qualitative Comparison}.} 
Fig.\ref{Qualitative} provides a visual comparison of the translation outputs of Qwen3-VL 8B, InternVL3, and GLoTran. 
Translated texts are overlaid on the original images using a rule-based mapping strategy across challenging scenarios, 
including complex visual backgrounds, dense document, curved text layout, and handwritten menu. 
Existing MLLMs frequently exhibit omissions, mistranslations, and hallucinations in these complex settings. 
For example, in menus and posters with small, scattered, or cluttered text, Qwen3-VL and InternVL3 fail to capture subtle textual cues, 
misrepresenting or omitting critical elements, whereas GLoTran consistently preserves textual fidelity and layout coherence.

\begin{table}[tbp]
\centering
\caption{Ablation study on the effect of GLoTran across different backbone models. 
We report \textbf{BLEU} and \textbf{COMET} scores for each translation task. 
The Gain row denotes the relative improvement (\%) of GLoTran over its baseline, 
calculated as $(\text{GLoTran} - \text{Baseline})/\text{Baseline} \times 100\%$.}
\renewcommand{\arraystretch}{0.9} 
\resizebox{\textwidth}{!}{
\begin{tabular}{c|cccccccc}
\toprule 
\textbf{Method} &  \textbf{zh $\rightarrow$ en}  & \textbf{zh $\rightarrow$ ko}   & \textbf{ko $\rightarrow$  zh}  & \textbf{jp $\rightarrow$ zh}  &  \textbf{Novel}  &  \textbf{Title}  &   \textbf{Introduction}  &  \textbf{Leaflet}   \\ \midrule
InternVL2.5 4B   & 4.02 / 70.78 &  1.19 / 60.44 & 2.57 / 48.20 &  28.37 / 76.66 & 18.48 / 75.67 & 31.44 / 76.10 & 26.16 / 71.09 &  21.22 / 63.92  \\ 
\rowcolor{gray!20}  GLoTran (InternVL2.5 4B) & 6.68 / 71.46 &  2.26 / 60.86 & 5.97 / 58.89 &  29.93 / 78.33 &  25.80 / 77.51 & 38.20 / 79.82 &  29.47 / 71.52 & 23.27 / 65.50  \\ 
\rowcolor{red!20} Gain \textcolor{red}{$\bigtriangleup$} &  +66.17 / +0.96 & +89.92 / +0.69 & +132.29 / +22.18 & +5.50 / +2.18 & +39.61 / +2.43 & +21.56 / +4.88 & +12.65 / +0.61 & +9.66 / +2.47    \\   \midrule
Qwen2.5-VL 7B & 9.54 / 74.45 & 4.13 / 66.76 & 20.23 / 70.27 & 35.83 / 79.09 &  27.43 / 78.09 &  47.91 / 80.09  &  36.18 / 72.78 &  43.66 / 72.40  \\
\rowcolor{gray!20}  GLoTran (Qwen2.5-VL 7B) &  10.99 / 76.78 & 5.45 / 68.49 & 22.39 /  70.99 &  43.81 / 81.84 & 44.59 / 83.60 &  53.15 / 82.92 &  38.93 / 74.12 &  45.64 / 73.74\\  
\rowcolor{red!20} Gain \textcolor{red}{$\bigtriangleup$} &  +15.20 / +3.13 & +31.96 / +2.59 & +10.68 / +1.02 & +22.27 / +3.48 & +62.59 / +7.06 & +10.93 / +3.53 & +7.60 / +1.84 & +4.54 / +1.85 \\    \midrule
Qwen3-VL 8B &  11.27 / 76.57 &  8.24 / 72.56 &  17.29 / 67.09 & 37.71 / 77.73 & 43.56 / 83.11 & 52.30 / 81.80 & 33.63 / 75.84 & 48.21 / 75.08 \\
\rowcolor{gray!20}  GLoTran (Qwen3-VL 8B)  & 11.87 /  79.27 &  9.62 / 75.46 & 20.69 / 69.73 &  43.54 / 81.92 & 46.84 / 86.61 & 54.33 / 83.21  & 41.32 / 76.40 & 51.18 / 77.91 \\
\rowcolor{red!20} Gain \textcolor{red}{$\bigtriangleup$} &   +5.33 / +3.53 & +16.75 / +4.00 & +19.66 / +3.94 & +15.46 / +5.39 & +7.53 / +4.21 & +3.88 / +1.72 & +22.86 / +0.74 & +6.16 / +3.77 \\ 
\bottomrule
\end{tabular}
}
\label{ab_backbone}
\end{table}

\subsection{Ablation Studies}
\textbf{Ablation in Backbone.}  We evaluate the effectiveness of GLoTran with different backbone models, 
including InternVL2.5 4B, Qwen2.5-VL 7B and Qwen3-VL 8B, as summarized in Tab.\ref{ab_backbone}.
Across all backbones, integrating GLoTran consistently yields substantial improvements in both BLEU and COMET scores.
For InternVL2.5 4B, GLoTran achieves notable gains, particularly on low-resource translation tasks such as ko$\rightarrow$zh and jp$\rightarrow$zh, with BLEU improvements of 132.3\% and 5.50\%, respectively.
Across other backbones, relative gains range from 4.54\% to 141.0\% compared to Qwen2.5-VL 7B, 
and up to 22.86\% BLEU in the Introduction and 6.16\% in the leaflet for Qwen3-VL 8B.
These results reveal two key insights: (\emph{i}) the global-local dual-perception strategy consistently improves translation 
fidelity across diverse backbones, and (\emph{ii}) its benefits are particularly pronounced for small-scale or visually intricate text regions,
highlighting that coordinated global-local perception yields greater gains than mere parameter scaling.

\textbf{Components analysis.} We employ the same text detector as GLoTran to extract text regions and feed these regions into existing MLLMs
to evaluate their translation performance under accurate text localization.
As shown in Tab.\ref{detector_only}, even when provided with localized text regions, the evaluated MLLMs 
(InternVL3$^\diamondsuit$, Qwen2.5-VL$^\diamondsuit$, and Qwen3-VL$^\diamondsuit$) 
still perform unsatisfactorily, with accuracy further declining in complex scenes, e.g., News and Sign.
To further isolate the effect of the text detector, we replace it in GLoTran  with a parameter-free shape-adaptive cropping (PSC) \cite{ye2023ureader,hu2025mplug}
strategy that partitions each high-resolution image into fixed-size local sub-images.
As reported in Tab.\ref{detector_only}, GLoTran (PSC) exhibits a significant performance decline 
due to the non-grounded random cropping strategy, a common practice adopted by existing MLLMs for high-resolution inputs, 
which introduces non-text regions, fragments text, and overlooks small text instances.
Moreover, we conduct an ablation by removing the global image and retaining only local slices.
As shown in Tab.\ref{detector_only}, GLoTran (w/o Global Image) exhibits a  performance drop compared to GLoTran,
onfirming the critical role of global contextual information.

\begin{table}[t]
\centering
\caption{Components analysis results (BLEU/COMET) of GLoTran on the MCiTon dataset.}
\resizebox{\columnwidth}{!}{
\begin{tabular}{l|c|ccc|ccc|cc}
\toprule 
&  & \multicolumn{3}{c|}{\textbf{Document}} & \multicolumn{3}{c|}{\textbf{Scene}} & \multicolumn{2}{c}{\textbf{Poster}} \\ \cmidrule(lr){3-5} \cmidrule(lr){6-8} \cmidrule(lr){9-10}
\multirow{-2}{*}{\textbf{Method}} & \multirow{-2}{*}{\textbf{Size}} & \textbf{Paper} $\uparrow$ & \textbf{News} $\uparrow$& \textbf{Novel} $\uparrow$ & \textbf{Title} $\uparrow$ & \textbf{Sign} $\uparrow$ & \textbf{Introduction} $\uparrow$ & \textbf{Cover} $\uparrow$ & \textbf{Leaflet} $\uparrow$  \\
\midrule 
InternVL3 &  8B   & 59.8/86.0 &  48.9/83.9  &  31.2/79.8 &  50.5/82.1 &  41.8/81.6 &  32.8/72.6 &  29.5/66.1  & 35.8/70.9 \\ 
InternVL3$^\diamondsuit$  &  8B   & 60.5/86.0 &  45.0/83.7  &  36.8/81.2 &  48.1/77.9 &  36.8/75.6 &  37.4/73.4 &  27.8/63.7  & 46.9/72.9 \\
Qwen2.5-VL &   7B   & 60.7/85.8 &  53.4/85.5  &  27.4/78.1 &  47.9/80.1 &  44.0/82.2 &  36.2/72.8 &  30.3/64.4  & 43.6/72.4 \\
Qwen2.5-VL$^\diamondsuit$  &  7B  & 63.5/86.5 &  46.3/84.0  &  37.6/81.9 &  42.6/78.8 &  37.5/75.2 &  40.3/73.4 &  32.9/65.8  & 44.6/72.9 \\
Qwen3-VL &   8B   & 60.9/86.1 &  54.3/86.2  &  43.6/83.1 &  52.3/81.8 &  44.2/83.0 &  33.6/75.8 &  36.2/70.4  & 48.2/75.1 \\
Qwen3-VL$^\diamondsuit$  &  8B    & 60.6/85.8 &  53.1/84.4  &  42.5/82.5 &  53.5/82.5 &  37.4/76.4 &  38.2/75.1 &  31.8/67.1  & 45.4/73.9 \\ 
\midrule
GLoTran (PSC)               & Qwen3-VL 8B  & 57.1/83.4 & 50.8/78.1 &  36.8/78.4 & 46.2/80.3 & 36.2/75.3 & 33.2/73.3 &  30.1/61.7 & 43.8/72.7  \\
GLoTran (w/o Global Image)  & Qwen3-VL 8B  & 60.7/85.9 & 44.8/83.7 & 44.1/82.9 & 41.0/75.6 & 37.8/77.4 & 33.1/74.6 &  34.1/67.6 & 44.9/73.1   \\
GLoTran (w/o Replay Mechanism) & Qwen3-VL 8B  & 64.8/85.9 & 56.0/86.7 & 45.5/84.8 & 53.1/81.2 & 45.0/82.7 & 40.0/74.5 & 36.1/69.9 & 49.5/75.8  \\  
GLoTran                     & Qwen3-VL 8B  &  \textbf{66.3/87.9} &  \textbf{57.5/88.6}  & \textbf{46.8/86.6} &  \textbf{54.3/83.2}  &   \textbf{46.3/84.8}  & \textbf{41.3/76.4} &  \textbf{37.7/72.0}  & \textbf{51.1/77.9}  \\
\bottomrule 
\end{tabular}
}
\label{detector_only}
\end{table}

\textbf{Hyperparameter Sensitivity Analysis.} 
Fig.\ref{multilingual_results}(c) shows the impact of varying global image resolutions (224, 448, 896, and 1792) on performance.
The results indicate that moderate resolutions (224 or 448) have the best accuracy.
Beyond 448, performance on introduction and leaflet scenes noticeably declines, likely due to excessive visual clutter and dispersed attention, which hinders effective global-context grounding.
Consequently, we adopt 224 as the default global resolution and complement it with local slices to retain fine-grained textual details.
Fig.\ref{multilingual_results}(d) presents the effect of the pre-region translation replay window $\eta$ for GLoTran (Qwen2.5-VL 7B).
BLEU peaks at 43.54 when $\eta=4$, while smaller or larger window sizes lead to performance degradation.
COMET scores fluctuate less but also favor moderate settings ($\eta \in [3,5]$) to achieve coherent translations.
An overly small $\eta$ fails to provide sufficient target-language context, resulting in local ambiguity and inconsistency.
Conversely, an excessively large $\eta$ introduces accumulated noise and translation drift, degrading contextual fidelity and increasing computational overhead.
In general, these sensitivity analyzes demonstrate that a balanced configuration of global visual resolution and contextual replay window 
maximizes the translation benefits of global-local cooperative modeling.

\begin{table}[tbp]
\centering
\caption{Comparison of MLLMs in terms of efficiency and accuracy for TIMI. 
The superscripts $\clubsuit$, $\heartsuit$, and $\spadesuit$ denote input resolutions of 224$\times$224,  448$\times$448, and the original size, respectively.}
\resizebox{0.8\columnwidth}{!}{
\begin{tabular}{r|cc|ccc|ccc}
\toprule
\multirow{2}{*}{\textbf{Method}} & \multirow{2}{*}{\textbf{Size}}  &  \multirow{2}{*}{\textbf{Token$^V$}$\downarrow$} & \multicolumn{3}{c|}{Novel} & \multicolumn{3}{c}{zh$\rightarrow$jp} \\ 
& & 
& \textbf{FTL$_s$}$\downarrow$ & \textbf{FPS}$\uparrow$ & \textbf{BLEU}$\uparrow$ 
& \textbf{FTL$_s$}$\downarrow$ & \textbf{FPS}$\uparrow$ & \textbf{BLEU}$\uparrow$ \\ 
\midrule
Qwen2.5-VL$^\heartsuit$                        & 7B & 6.1K    & 0.098 & 0.061 & 27.36 & 0.098  & 3.755  & 11.85 \\
InternVL3$^\heartsuit$                         & 8B & 10K     & 0.033 & 0.065 & 31.74 & 0.149  & 2.016  & 15.02 \\ 
Qwen3-VL$^\heartsuit$                          & 8B & 6.1K    & 0.099 & 0.028 & 9.98  & 0.093  & 2.572  & 18.79 \\
Qwen3-VL$^\spadesuit$                          & 8B & 164K    & 1.081 & 0.041 & 43.56 & 3.297  & 0.275  & 22.06 \\
\rowcolor[HTML]{D6ECFF} GLoTran$^\heartsuit$   & 8B & 8.4K    & 3.002 & 0.035 & 48.12 & 0.253  & 2.285  & 26.74 \\
\rowcolor[HTML]{D6ECFF} GLoTran$^\clubsuit$    & 8B & 4.9K    & 2.975 & 0.037 & 46.84 & 0.241  & 2.303  & 25.53 \\
\bottomrule
\end{tabular}}
\label{efficiency_main}
\end{table}

\textbf{Complexity Analysis.}
To evaluate the trade-off between efficiency and accuracy, 
we conduct a complexity analysis of representative MLLMs on two challenging benchmarks, 
Novel  (all images<2K) and zh$\rightarrow$jp (130/200 images $\geqslant$ 2K and 53/200 images $\geqslant$ 4K), 
which contain text-rich and visually complex high-resolution images. 
Tab.\ref{efficiency_main} reports the average number of visual tokens per image ($\text{Token}^V$), 
First Token Latency (FTL$_s$), frame processing speed (FPS), and BLEU score. Under a unified resolution of 448$\times$448,
GLoTran$^\heartsuit$ uses more visual tokens than Qwen3-VL$^\heartsuit$ and Qwen2.5-VL$^\heartsuit$,
but achieves substantially higher BLEU scores, surpassing Qwen3-VL$^\heartsuit$ by 
38.14 and 7.95 points on Novel and zh$\rightarrow$jp, respectively. 
Achieving comparable accuracy with Qwen3-VL$^\spadesuit$ requires processing full-resolution images, 
which dramatically increases visual tokens (164K vs. 6.1K), first token latency (3.297 vs. 0.093),
and computational cost ($\thicksim$215$\times$ more FLOPs) in Novel.
In contrast, our global-local dual perception framework enables GLoTran$^\clubsuit$ to maintain competitive performance
even at lower resolutions 224$\times$224, capturing fine-grained textual details and global contextual layouts 
while keeping computational overhead manageable, requiring only 4.9K $\text{Token}^V$ and about 18GB of memory.

These results indicate that existing MLLMs rely heavily on resolution scaling to improve accuracy,
 which introduces substantial visual redundancy with limited gains, whereas our global-local dual perception framework 
 captures fine-grained text and global context even at low resolutions, enabling high-quality translation with lower computational cost.

\section{Conclusion}
\label{Conclusion}
In this paper, we propose GLoTran, a global-local dual visual perception framework for MLLMs in TIMT. 
By integrating global contextual understanding with fine-grained local text focus, 
GLoTran enables MLLMs to capture holistic semantics and precise localized details, 
thereby supporting accurate translation of high-resolution text-rich images.
To facilitate global-local dual-perception learning for existing MLLMs, we further construct GLoD, 
a large-scale dataset containing over 510K global-local 
image-text pairs, systematically curated through a unified data generation pipeline. 
Extensive experiments demonstrate that GLoTran achieves improvements over state-of-the-art MLLMs,
delivering superior translation completeness and reliability.

\section*{Acknowledgements}
This work is supported by the National Natural Science Foundation of China under Project 62576206. 

%
%
\bibliographystyle{splncs04}
\bibliography{main}

@String(CVPR  = {IEEE Conf. Comput. Vis. Pattern Recog.})

@String(NeurIPS = {Adv. Neural Inform. Process. Syst.})

@String(ICLR  = {Int. Conf. Learn. Represent.})

@String(AAAI  = {AAAI})

@String(IJCAI = {IJCAI})

@String(ICPR  = {Int. Conf. Pattern Recog.})

@String(CVPR  = {CVPR})

@String(NeurIPS = {NeurIPS})

@String(ICLR  = {ICLR})

@String(ICPR  = {ICPR})

@inproceedings{lan2024translatotron,
  title={Translatotron-V (ison): An End-to-End Model for In-Image Machine Translation},
  author={Lan, Zhibin and Niu, Liqiang and Meng, Fandong and Zhou, Jie and Zhang, Min and Su, Jinsong},
  booktitle={finding of ACL},
  pages={5472--5485},
  year={2024}
}

@inproceedings{yanglanguage,
  title={Language Imbalance Driven Rewarding for Multilingual Self-improving},
  author={Yang, Wen and Wu, Junhong and Wang, Chen and Zong, Chengqing and Zhang, Jiajun},
  booktitle={ICLR},
    year={2025}
}

@inproceedings{zhao2021knowledge,
  title={Knowledge graphs enhanced neural machine translation},
  author={Zhao, Yang and Zhang, Jiajun and Zhou, Yu and Zong, Chengqing},
  booktitle={IJCAI},
  pages={4039--4045},
  year={2021}
}

@article{castilho2025survey,
  title={A survey of context in neural machine translation and its evaluation},
  author={Castilho, Sheila and Knowles, Rebecca},
  journal={Nat. Lang. Process.},
  volume={31},
  number={4},
  pages={986--1016},
  year={2025}
}

@inproceedings{li2024non,
  title={Non-Autoregressive Machine Translation as Constrained HMM},
  author={Li, Haoran and Jie, Zhanming and Lu, Wei},
  booktitle={finding of ACL},
  pages={12361--12372},
  year={2024}
}

@inproceedings{mansimov2020towards,
  title={Towards End-to-End In-Image Neural Machine Translation},
  author={Mansimov, Elman and Stern, Mitchell and Chen, Mia Xu and Firat, Orhan and Uszkoreit, Jakob and Jain, Puneet},
  booktitle={NLPBW},
  pages={70--74},
  year={2020}
}

@inproceedings{zuo2025inimagetrans,
  title={InImageTrans: Multimodal LLM-based Text Image Machine Translation},
  author={Zuo, Fei and Chen, Kehai and Zhang, Yu and Xue, Zhengshan and Zhang, Min},
  booktitle={finding of ACL},
  pages={20256--20277},
  year={2025}
}

@inproceedings{rei2020comet,
  title={COMET: A Neural Framework for MT Evaluation},
  author={Rei, Ricardo and Stewart, Craig and Farinha, Ana C and Lavie, Alon},
  booktitle={EMNLP},
  pages={2685--2702},
  year={2020}
}

@inproceedings{papineni2002bleu,
  title={Bleu: a method for automatic evaluation of machine translation},
  author={Papineni, Kishore and Roukos, Salim and Ward, Todd and Zhu, Wei-Jing},
  booktitle={ACL},
  pages={311--318},
  year={2002}
}

@inproceedings{zhang2023novel,
  title={A novel dataset and benchmark analysis on document image translation},
  author={Zhang, Zhiyang and Zhang, Yaping and Xiang, Lu and Zhao, Yang and Zhou, Yu and Zong, Chengqing},
  booktitle={CCMT},
  pages={103--115},
  year={2023},
  organization={Springer}
}

@article{blecher2023nougat,
  title={Nougat: Neural optical understanding for academic documents},
  author={Blecher, Lukas and Cucurull, Guillem and Scialom, Thomas and Stojnic, Robert},
  journal={arXiv preprint arXiv:2308.13418},
  year={2023}
}

@inproceedings{hinami2021towards,
  title={Towards fully automated manga translation},
  author={Hinami, Ryota and Ishiwatari, Shonosuke and Yasuda, Kazuhiko and Matsui, Yusuke},
  booktitle={AAAI},
  volume={35},
  number={14},
  pages={12998--13008},
  year={2021}
}

@inproceedings{su2021rtnet,
  title={Rtnet: An end-to-end method for handwritten text image translation},
  author={Su, Tonghua and Liu, Shuchen and Zhou, Shengjie},
  booktitle={ICDAR},
  pages={99--113},
  year={2021},
  organization={Springer}
}

@article{ma2023modal,
  title={Modal contrastive learning based end-to-end text image machine translation},
  author={Ma, Cong and Han, Xu and Wu, Linghui and Zhang, Yaping and Zhao, Yang and Zhou, Yu and Zong, Chengqing},
  journal={ACM Trans. Audio Speech Lang. Process.},
  volume={32},
  pages={2153--2165},
  year={2023}
}

@inproceedings{ma2022improving,
  title={Improving end-to-end text image translation from the auxiliary text translation task},
  author={Ma, Cong and Zhang, Yaping and Tu, Mei and Han, Xu and Wu, Linghui and Zhao, Yang and Zhou, Yu},
  booktitle={ICPR},
  pages={1664--1670},
  year={2022},
  organization={IEEE}
}

@inproceedings{niu2024umtit,
  title={UMTIT: Unifying recognition, translation, and generation for multimodal text image translation},
  author={Niu, Liqiang and Meng, Fandong and Zhou, Jie},
  booktitle={LREC-COLING},
  pages={16953--16972},
  year={2024}
}

@article{gao2024towards,
  title={Towards boosting many-to-many multilingual machine translation with large language models},
  author={Gao, Pengzhi and He, Zhongjun and Wu, Hua and Wang, Haifeng},
  journal={arXiv preprint arXiv:2401.05861},
  year={2024}
}

@inproceedings{liang2025single,
  title={Single-to-mix Modality Alignment with Multimodal Large Language Model for Document Image Machine Translation},
  author={Liang, Yupu and Zhang, Yaping and Zhang, Zhiyang and Zhao, Yang and Xiang, Lu and Zong, Chengqing and Zhou, Yu},
  booktitle={ACL},
  pages={12391--12408},
  year={2025}
}

@inproceedings{liang2025improving,
  title={Improving MLLM’s Document Image Machine Translation via Synchronously Self-reviewing Its OCR Proficiency},
  author={Liang, Yupu and Zhang, Yaping and Zhang, Zhiyang and Chen, Zhiyuan and Zhao, Yang and Xiang, Lu and Zong, Chengqing and Zhou, Yu},
  booktitle={finding of ACL},
  pages={23659--23678},
  year={2025}
}

@inproceedings{qian2024anytrans,
  title={AnyTrans: Translate AnyText in the Image with Large Scale Models},
  author={Qian, Zhipeng and Zhang, Pei and Yang, Baosong and Fan, Kai and Ma, Yiwei and Wong, Derek and Sun, Xiaoshuai and Ji, Rongrong},
  booktitle={finding of EMNLP},
  pages={2432--2444},
  year={2024}
}

@inproceedings{rafailov2023direct,
    title={Direct Preference Optimization: Your Language Model is Secretly a Reward Model},
    author={Rafael Rafailov and Archit Sharma and Eric Mitchell and Christopher D Manning and Stefano Ermon and Chelsea Finn},
    booktitle={NeurIPS},
    year={2023},
    url={https://arxiv.org/abs/2305.18290}
}

@inproceedings{zhang2024paying,
  title={Paying More Attention to Source Context: Mitigating Unfaithful Translations from Large Language Model},
  author={Zhang, Hongbin and Chen, Kehai and Bai, Xuefeng and Xiang, Yang and Zhang, Min},
  booktitle={finding of ACL},
  year={2024}
}

@inproceedings{ma2023multi,
  title={Multi-teacher knowledge distillation for end-to-end text image machine translation},
  author={Ma, Cong and Zhang, Yaping and Tu, Mei and Zhao, Yang and Zhou, Yu and Zong, Chengqing},
  booktitle={ICDAR},
  pages={484--501},
  year={2023},
  organization={Springer}
}

@inproceedings{park2024hierarchical,
  title={Hierarchical visual feature aggregation for ocr-free document understanding},
  author={Park, Jaeyoo and Choi, Jin Y and Park, Jeonghyung and Han, Bohyung},
  booktitle={NeurIPS},
  volume={37},
  pages={105972--105996},
  year={2024}
}

@inproceedings{fang2025recognition,
  title={Recognition-Synergistic Scene Text Editing},
  author={Fang, Zhengyao and Lyu, Pengyuan and Wu, Jingjing and Zhang, Chengquan and Yu, Jun and Lu, Guangming and Pei, Wenjie},
  booktitle={CVPR},
  pages={13104--13113},
  year={2025}
}

@inproceedings{ye2023ureader,
  title={UReader: Universal OCR-free Visually-situated Language Understanding with Multimodal Large Language Model},
  author={Ye, Jiabo and Hu, Anwen and Xu, Haiyang and Ye, Qinghao and Yan, Ming and Xu, Guohai and Li, Chenliang and Tian, Junfeng and Qian, Qi and Zhang, Ji and others},
  booktitle={finding of EMNLP},
  pages={2841--2858},
  year={2023}
}

@article{vu2025describe,
  title={Describe Anything Model for Visual Question Answering on Text-rich Images},
  author={Vu, Yen-Linh and Duong, Dinh-Thang and Duong, Truong-Binh and Nguyen, Anh-Khoi and Nguyen, Thanh-Huy and Nguyen, Le Thien Phuc and Xing, Jianhua and Li, Xingjian and Wang, Tianyang and Bagci, Ulas and others},
  journal={arXiv preprint arXiv:2507.12441},
  year={2025}
}

@article{luan2024textcot,
  title={Textcot: Zoom in for enhanced multimodal text-rich image understanding},
  author={Luan, Bozhi and Feng, Hao and Chen, Hong and Wang, Yonghui and Zhou, Wengang and Li, Houqiang},
  journal={arXiv preprint arXiv:2404.09797},
  year={2024}
}

@inproceedings{zhang2025chaotic,
  title={From chaotic ocr words to coherent document: A fine-to-coarse zoom-out network for complex-layout document image translation},
  author={Zhang, Zhiyang and Zhang, Yaping and Liang, Yupu and Xiang, Lu and Zhao, Yang and Zhou, Yu and Zong, Chengqing},
  booktitle={COLING},
  pages={10877--10890},
  year={2025}
}

@inproceedings{fang2025guided,
  title={guided MLLM Reasoning: Enhancing MLLM with Knowledge and Visual Notes for Visual Question Answering},
  author={Fang, Wenlong and Wu, Qiaofeng and Chen, Jing and Xue, Yun},
  booktitle={CVPR},
  pages={19597--19607},
  year={2025}
}

@inproceedings{zhao2024lova3,
  title={Lova3: Learning to visual question answering, asking and assessment},
  author={Zhao, Henry Hengyuan and Zhou, Pan and Gao, Difei and Bai, Zechen and Shou, Mike Zheng},
  booktitle={NeurIPS},
  volume={37},
  pages={115146--115175},
  year={2024}
}

@inproceedings{tian2023image,
  title={In-image neural machine translation with segmented pixel sequence-to-sequence model},
  author={Tian, Yanzhi and Li, Xiang and Liu, Zeming and Guo, Yuhang and Wang, Bin},
  booktitle={finding of EMNLP },
  pages={15046--15057},
  year={2023}
}

@inproceedings{zhang2023layoutdit,
  title={LayoutDIT: Layout-aware end-to-end document image translation with multi-step conductive decoder},
  author={Zhang, Zhiyang and Zhang, Yaping and Liang, Yupu and Xiang, Lu and Zhao, Yang and Zhou, Yu and Zong, Chengqing},
  booktitle={finding of EMNLP},
  pages={10043--10053},
  year={2023}
}

@inproceedings{gao2025multimodal,
  title={Multimodal Machine Translation with Text-Image In-depth Questioning},
  author={Gao, Yue and Zhao, Jing and Sun, Shiliang and Qiao, Xiaosong and Song, Tengfei and Yang, Hao},
  booktitle={finding of ACL},
  pages={9274--9287},
  year={2025}
}

@article{cui2025paddleocr,
  title={Paddleocr 3.0 technical report},
  author={Cui, Cheng and Sun, Ting and Lin, Manhui and Gao, Tingquan and Zhang, Yubo and Liu, Jiaxuan and Wang, Xueqing and Zhang, Zelun and Zhou, Changda and Liu, Hongen and others},
  journal={arXiv preprint arXiv:2507.05595},
  year={2025}
}

@article{dosovitskiy2020image,
  title={An image is worth 16x16 words: Transformers for image recognition at scale},
  author={Dosovitskiy, Alexey},
  journal={arXiv preprint arXiv:2010.11929},
  year={2020}
}

@article{dong2024internlm,
  title={Internlm-xcomposer2: Mastering free-form text-image composition and comprehension in vision-language large model},
  author={Dong, Xiaoyi and Zhang, Pan and Zang, Yuhang and Cao, Yuhang and Wang, Bin and Ouyang, Linke and Wei, Xilin and Zhang, Songyang and Duan, Haodong and Cao, Maosong and others},
  journal={arXiv preprint arXiv:2401.16420},
  year={2024}
}

@inproceedings{wang2024cogvlm,
  title={CogVLM: visual expert for pretrained language models},
  author={Wang, Weihan and Lv, Qingsong and Yu, Wenmeng and Hong, Wenyi and Qi, Ji and Wang, Yan and Ji, Junhui and Yang, Zhuoyi and Zhao, Lei and Song, Xixuan and others},
  booktitle={NeurIPS},
  pages={121475--121499},
  year={2024}
}

@inproceedings{chen2024internvl,
  title={Internvl: Scaling up vision foundation models and aligning for generic visual-linguistic tasks},
  author={Chen, Zhe and Wu, Jiannan and Wang, Wenhai and Su, Weijie and Chen, Guo and Xing, Sen and Zhong, Muyan and Zhang, Qinglong and Zhu, Xizhou and Lu, Lewei and others},
  booktitle={CVPR},
  pages={24185--24198},
  year={2024}
}

@article{zhu2025internvl3,
  title={Internvl3: Exploring advanced training and test-time recipes for open-source multimodal models},
  author={Zhu, Jinguo and Wang, Weiyun and Chen, Zhe and Liu, Zhaoyang and Ye, Shenglong and Gu, Lixin and Tian, Hao and Duan, Yuchen and Su, Weijie and Shao, Jie and others},
  journal={arXiv preprint arXiv:2504.10479},
  year={2025}
}

@article{wang2024qwen2,
  title={Qwen2-vl: Enhancing vision-language model's perception of the world at any resolution},
  author={Wang, Peng and Bai, Shuai and Tan, Sinan and Wang, Shijie and Fan, Zhihao and Bai, Jinze and Chen, Keqin and Liu, Xuejing and Wang, Jialin and Ge, Wenbin and others},
  journal={arXiv preprint arXiv:2409.12191},
  year={2024}
}

@article{bai2025qwen2,
  title={Qwen2.5-vl technical report},
  author={Bai, Shuai and Chen, Keqin and Liu, Xuejing and Wang, Jialin and Ge, Wenbin and Song, Sibo and Dang, Kai and Wang, Peng and Wang, Shijie and Tang, Jun and others},
  journal={arXiv preprint arXiv:2502.13923},
  year={2025}
}

@article{yang2025qwen3,
  title={Qwen3 technical report},
  author={Yang, An and Li, Anfeng and Yang, Baosong and Zhang, Beichen and Hui, Binyuan and Zheng, Bo and Yu, Bowen and Gao, Chang and Huang, Chengen and Lv, Chenxu and others},
  journal={arXiv preprint arXiv:2505.09388},
  year={2025}
}

@article{bai2023qwen,
  title={Qwen-vl: A frontier large vision-language model with versatile abilities},
  author={Bai, Jinze and Bai, Shuai and Yang, Shusheng and Wang, Shijie and Tan, Sinan and Wang, Peng and Lin, Junyang and Zhou, Chang and Zhou, Jingren},
  journal={arXiv preprint arXiv:2308.12966},
  volume={1},
  number={2},
  pages={3},
  year={2023}
}

@inproceedings{li202510m,
  title={MIT-10M: A Large Scale Parallel Corpus of Multilingual Image Translation},
  author={Li, Bo and Zhu, Shaolin and Wen, Lijie},
  booktitle={COLING},
  pages={5154--5167},
  year={2025}
}

@inproceedings{liang2024document,
  title={Document image machine translation with dynamic multi-pre-trained models assembling},
  author={Liang, Yupu and Zhang, Yaping and Ma, Cong and Zhang, Zhiyang and Zhao, Yang and Xiang, Lu and Zong, Chengqing and Zhou, Yu},
  booktitle={NAACL},
  pages={7084--7095},
  year={2024}
}

@inproceedings{lan2023exploring,
  title={Exploring Better Text Image Translation with Multimodal Codebook},
  author={Lan, Zhibin and Yu, Jiawei and Li, Xiang and Zhang, Wen and Luan, Jian and Wang, Bin and Huang, Degen and Su, Jinsong},
  booktitle={ACL},
  pages={3479--3491},
  year={2023}
}

@inproceedings{zhu2023peit,
  title={PEIT: bridging the modality gap with pre-trained models for end-to-end image translation},
  author={Zhu, Shaolin and Li, Shangjie and Lei, Yikun and Xiong, Deyi},
  booktitle={ACL},
  pages={13433--13447},
  year={2023}
}

@article{guo2025deepseek,
  title={Deepseek-r1: Incentivizing reasoning capability in llms via reinforcement learning},
  author={Guo, Daya and Yang, Dejian and Zhang, Haowei and Song, Junxiao and Zhang, Ruoyu and Xu, Runxin and Zhu, Qihao and Ma, Shirong and Wang, Peiyi and Bi, Xiao and others},
  journal={arXiv preprint arXiv:2501.12948},
  year={2025}
}

@article{hurst2024gpt,
  title={{GPT-4o} system card},
  author={Hurst, Aaron and Lerer, Adam and Goucher, Adam P and Perelman, Adam and Ramesh, Aditya and Clark, Aidan and Ostrow, AJ and Welihinda, Akila and Hayes, Alan and Radford, Alec and others},
  journal={arXiv preprint arXiv:2410.21276},
  year={2024}
}

@inproceedings{gao2021simcse,
  title={SimCSE: Simple Contrastive Learning of Sentence Embeddings},
  author={Gao, Tianyu and Yao, Xingcheng and Chen, Danqi},
  booktitle={EMNLP},
  pages={6894--6910},
  year={2021}
}

@article{feng2020language,
  title={Language-agnostic BERT sentence embedding},
  author={Feng, Fangxiaoyu and Yang, Yinfei and Cer, Daniel and Arivazhagan, Naveen and Wang, Wei},
  journal={arXiv preprint arXiv:2007.01852},
  year={2020}
}

@article{dong2025scalable,
  title={Scalable vision language model training via high quality data curation},
  author={Dong, Hongyuan and Kang, Zijian and Yin, Weijie and Liang, Xiao and Feng, Chao and Ran, Jiao},
  journal={arXiv preprint arXiv:2501.05952},
  year={2025}
}

@article{dong2025qianfan,
  title={Qianfan-vl: Domain-enhanced universal vision-language models},
  author={Dong, Daxiang and Zheng, Mingming and Xu, Dong and Zhuang, Bairong and Zhang, Wenyu and Luo, Chunhua and Wang, Haoran and Zhao, Zijian and Li, Jie and Li, Yuxuan and others},
  journal={arXiv preprint arXiv:2509.18189},
  year={2025}
}

@article{lu2024deepseek,
  title={Deepseek-vl: towards real-world vision-language understanding},
  author={Lu, Haoyu and Liu, Wen and Zhang, Bo and Wang, Bingxuan and Dong, Kai and Liu, Bo and Sun, Jingxiang and Ren, Tongzheng and Li, Zhuoshu and Yang, Hao and others},
  journal={arXiv preprint arXiv:2403.05525},
  year={2024}
}

@article{feng2025mt,
  title={MT3: Scaling MLLM-based Text Image Machine Translation via Multi-Task Reinforcement Learning},
  author={Feng, Zhaopeng and Liang, Yupu and Cao, Shaosheng and Su, Jiayuan and Ren, Jiahan and Xu, Zhe and Hu, Yao and Huang, Wenxuan and Wu, Jian and Liu, Zuozhu},
  journal={arXiv preprint arXiv:2505.19714},
  year={2025}
}

@article{chen2024diffusion,
  title={Diffusion forcing: Next-token prediction meets full-sequence diffusion},
  author={Chen, Boyuan and Mart{\'\i} Mons{\'o}, Diego and Du, Yilun and Simchowitz, Max and Tedrake, Russ and Sitzmann, Vincent},
  journal={NIPS},
  volume={37},
  pages={24081--24125},
  year={2024}
}

@inproceedings{tian2025beyond,
  title={Beyond answers: Transferring reasoning capabilities to smaller llms using multi-teacher knowledge distillation},
  author={Tian, Yijun and Han, Yikun and Chen, Xiusi and Wang, Wei and Chawla, Nitesh V},
  booktitle={WSDM},
  pages={251--260},
  year={2025}
}

@article{wu2025flashbias,
  title={FlashBias: Fast Computation of Attention with Bias},
  author={Wu, Haixu and Guo, Minghao and Ma, Yuezhou and Sun, Yuanxu and Wang, Jianmin and Matusik, Wojciech and Long, Mingsheng},
  journal={arXiv preprint arXiv:2505.12044},
  year={2025}
}

@inproceedings{yemplug,
  title={mPLUG-Owl3: Towards Long Image-Sequence Understanding in Multi-Modal Large Language Models},
  author={Ye, Jiabo and Xu, Haiyang and Liu, Haowei and Hu, Anwen and Yan, Ming and Qian, Qi and Zhang, Ji and Huang, Fei and Zhou, Jingren},
  booktitle={ICLR}
}

@inproceedings{hu2025mplug,
  title={mplug-docowl2: High-resolution compressing for ocr-free multi-page document understanding},
  author={Hu, Anwen and Xu, Haiyang and Zhang, Liang and Ye, Jiabo and Yan, Ming and Zhang, Ji and Jin, Qin and Huang, Fei and Zhou, Jingren},
  booktitle={Proceedings of the 63rd Annual Meeting of the Association for Computational Linguistics (Volume 1: Long Papers)},
  pages={5817--5834},
  year={2025}
}

@article{wang2024mineru,
  title={Mineru: An open-source solution for precise document content extraction},
  author={Wang, Bin and Xu, Chao and Zhao, Xiaomeng and Ouyang, Linke and Wu, Fan and Zhao, Zhiyuan and Xu, Rui and Liu, Kaiwen and Qu, Yuan and Shang, Fukai and others},
  journal={arXiv preprint arXiv:2409.18839},
  year={2024}
}

@inproceedings{wu2024v,
  title={V*: Guided visual search as a core mechanism in multimodal llms},
  author={Wu, Penghao and Xie, Saining},
  booktitle={Proceedings of the IEEE/CVF Conference on Computer Vision and Pattern Recognition},
  pages={13084--13094},
  year={2024}
}

@inproceedings{wang2025divide,
  title={Divide, conquer and combine: A training-free framework for high-resolution image perception in multimodal large language models},
  author={Wang, Wenbin and Ding, Liang and Zeng, Minyan and Zhou, Xiabin and Shen, Li and Luo, Yong and Yu, Wei and Tao, Dacheng},
  booktitle={Proceedings of the AAAI Conference on Artificial Intelligence},
  volume={39},
  number={8},
  pages={7907--7915},
  year={2025}
}

@inproceedings{shen2025zoomeye,
  title={Zoomeye: Enhancing multimodal llms with human-like zooming capabilities through tree-based image exploration},
  author={Shen, Haozhan and Zhao, Kangjia and Zhao, Tiancheng and Xu, Ruochen and Zhang, Zilun and Zhu, Mingwei and Yin, Jianwei},
  booktitle={Proceedings of the 2025 Conference on Empirical Methods in Natural Language Processing},
  pages={6613--6629},
  year={2025}
}

@inproceedings{cailocal,
  title={From Local Details to Global Context: Advancing Vision-Language Models with Attention-Based Selection},
  author={Cai, Lincan and Kang, Jingxuan and Li, Shuang and Ma, Wenxuan and Xie, Binhui and Qin, Zhida and Liang, Jian},
  booktitle={Forty-second International Conference on Machine Learning}
}
\end{document}